\documentclass[10pt,twocolumn,letterpaper]{article}

\usepackage{arydshln}
\usepackage{iccv}
\usepackage{times}
\usepackage{epsfig}
\usepackage{graphicx}
\usepackage{amsmath}
\usepackage{amssymb}
\usepackage{multirow}
\usepackage{booktabs}
\usepackage{comment}
\usepackage[accsupp]{axessibility} 

\usepackage[breaklinks=true,bookmarks=false]{hyperref}

\iccvfinalcopy 

\def\iccvPaperID{****} 

\ificcvfinal\fi

\begin{document}

\title{Modelling Neighbor Relation in Joint Space-Time Graph\\ for Video Correspondence Learning}

\author{Zixu Zhao \qquad Yueming Jin \qquad Pheng-Ann Heng\\
The Chinese University of Hong Kong \\
{\tt\small \{zxzhao, ymjin, pheng\}@cse.cuhk.edu.hk}}

\maketitle
\ificcvfinal\thispagestyle{empty}\fi

\begin{abstract}
   This paper presents a self-supervised method for learning reliable visual correspondence from unlabeled videos. We formulate the correspondence as finding paths in a joint space-time graph, where nodes are grid patches sampled from frames, and are linked by two type of edges: (i) neighbor relations that determine the aggregation strength from  intra-frame neighbors in space, and (ii) similarity relations that indicate the transition probability of inter-frame paths across time. Leveraging the cycle-consistency in videos, our contrastive learning objective discriminates dynamic objects from both their neighboring views and temporal views. Compared with prior works, our approach actively explores the neighbor relations of central instances to learn a latent association between center-neighbor pairs (\eg, ``hand -- arm'') across time, thus improving the instance discrimination. Without fine-tuning, our learned representation outperforms the state-of-the-art self-supervised methods on a variety of visual  tasks including video object propagation, part propagation, and pose keypoint tracking. Our self-supervised method also surpasses some fully supervised algorithms designed for the specific tasks. 
\end{abstract}

\section{Introduction}

Learning temporal correspondence --- a problem of learning ``what went where''--- is closely related to many fundamental vision tasks, such as video object tracking~\cite{wang2019unsupervised, liu2010sift, vondrick2018tracking}, video object segmentation~\cite{wang2019fast, caelles2017one, voigtlaender2017online, maninis2018video, oh2019video}, and flow estimation~\cite{dosovitskiy2015flownet, ilg2017flownet}. 
In essence, it corresponds to a query-target matching problem, which relies on an affinity to match a physical point (or patch) in the query frame $t$ to that in the target frame $t+k$.
One practical issue is collecting dense annotations from large-scale videos, which costs large human efforts. It motivates numerous self-supervised methods ~\cite{vondrick2018tracking, wang2019learning, li2019joint, lai2019self, lai2020mast, Wang_2021_Contrastive, jabri2020walk} to learn dynamic objects from unlabeled videos by leveraging the cycle-consistency in time as a free supervisory signal.

\begin{figure}
	\centering
	\includegraphics[width=85mm]{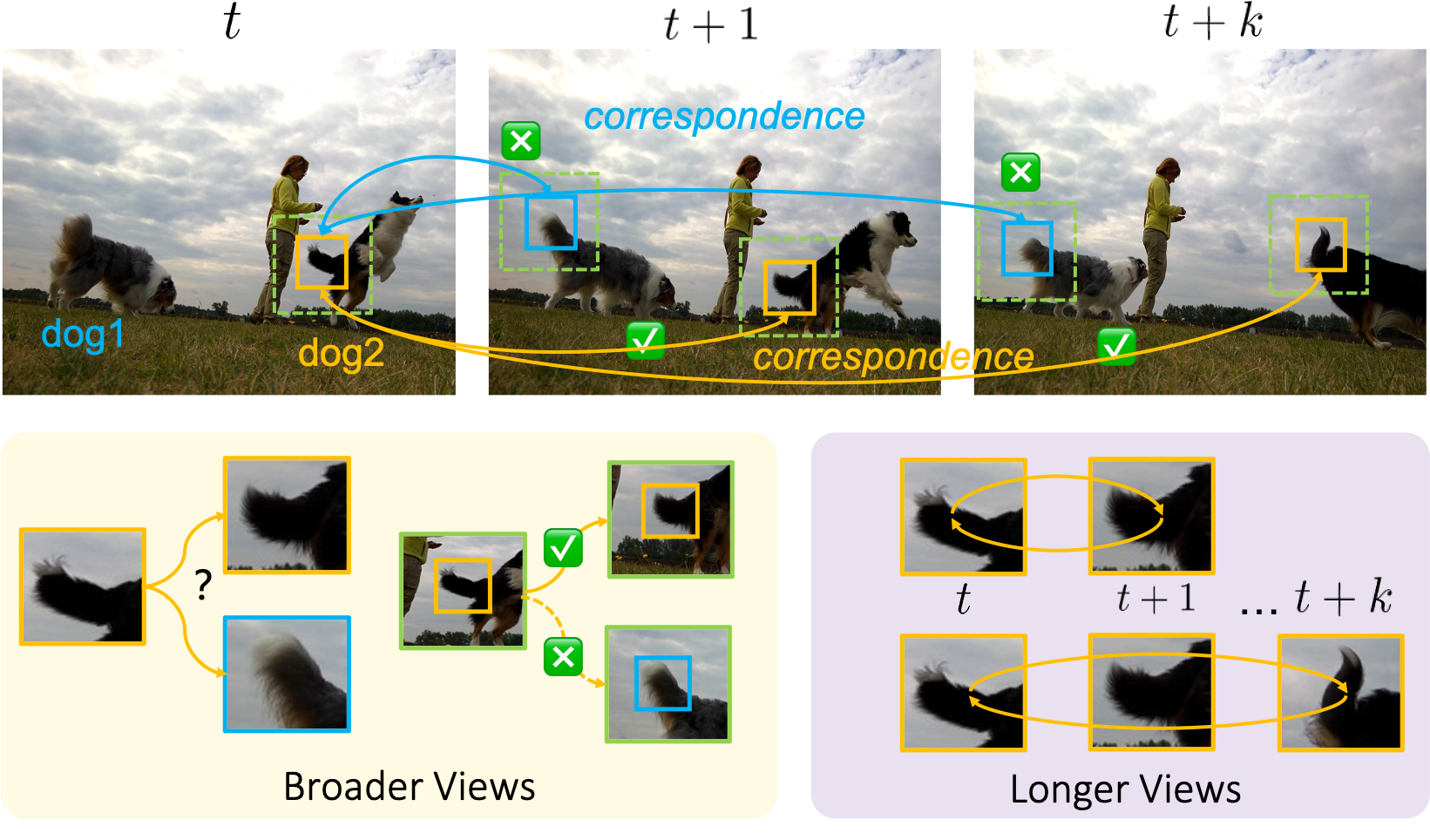}
\caption{How to find the correspondence of a small object such as the dog tail in a video? We argue that the query-target matching desires both \textit{longer views} (temporal dynamics) and \textit{broader views} (neighbor relations) to distinguish similar instances.
We capture these two cues in a graph to learn correspondence.}
\label{fig:intro}
\vspace{-3mm}
\end{figure}

Recent approaches learn strong representations by constructing long-range views mainly from two perspectives: (i) learning pixel-level correspondences by a \textit{single-step} association~\cite{li2019joint, Wang_2021_Contrastive}, 
or (ii) learning patch-level correspondences by a \textit{multi-step} association~\cite{jabri2020walk}. The single-step association can be viewed as a pixel-level affinity between two patches at timesteps $t$ and $t+k$, aiming to transform the pixel colors. 
Such transformation requires a deterministic correspondence to locate the target patch at $t+k$, which is achieved by training an extra unsupervised patch tracker~\cite{wang2019learning}. However, the underlying assumption that corresponding pixels have the same color may be violated, \eg, inevitable lighting changes and deformation in future frames, thereby hindering the model from using longer temporal cues. 
Recently, Jabri \etal~\cite{jabri2020walk} formulate a multi-step association that connects 
corresponding patches at every timestep between $t$ and $t+k$ in a form of Markov chain. At each step, the patch-level affinity links all patches of two adjacent frames, preserving all possible correspondences in the video.
The learning process thus benefits from a longer-range clip with all intermediate views available.
Unfortunately, finding an ``optimal'' correspondence is not easy due to the struggling matching between similar instances where image patches only capture a very \textit{narrow} view of them.

Hence, we identify another key ingredient for  better query-target matching --- \textit{seeing broader} --- which is ignored in existing methods.
Let us take a closer look of an example shown in Figure~\ref{fig:intro}. 
How do human track the right dog tail across frames, and avoid being confused by the  left dog tail from the similar instance? 
(i) \textit{Seeing longer}: how the shape of the tail changes over time is indeed a crucial cue, and it can be utilized in the multi-step association~\cite{jabri2020walk}.
(ii) \textit{Seeing broader}: it is easier to discriminate the dog tails from a broader view by considering neighboring information around them, such as the dog body features, as well as the dog-person interaction. However, it cannot be achieved by straightforwardly enlarging the patch size, as the detailed structures or features will be missed.

In this paper, we propose to learn correspondence by \textit{seeing both broader and longer} via a graph-based framework.
We represent the video as a joint space-time graph where
nodes are grid patches and edges are two type of relations, \ie, neighbor relations and similarity relations. The big graph can thus be decomposed into two sub-graphs.
(i) \textbf{Neighbor Relation Graph}: we start by constructing a small graph for each node, which is  linked to intra-frame nodes that are located in a sliding neighborhood. Initialized with the topological prior, the edges learn to guide the aggregation of neighboring node representations to the central node.
The updated node representation thus captures a broader view of the neighborhood.
(ii) \textbf{Similarity Graph}: we then connect inter-frame nodes with the pairwise similarity, under the updated node representations. All these edges form a multi-step association for a long-range clip.

Given the joint graph, the prediction of the long-range correspondence can be computed as a path (a combination of similarity-based edges) along the graph.
To induce supervision, we adopt palindrome sequences~\cite{jabri2020walk} for training, which provide the walker with a target, \ie, returning to the start point. In contrast to the prior work~\cite{jabri2020walk}, our path-level constraint provides contrastive learning signals from both the temporal views and neighboring views, leading to more reliable matching among similar instances. Moreover, we perform a random but attentive walk on the large graph by wisely dropping the ``common-fate''~\cite{wertheimer1938laws} nodes according to the pixel discrepancy of each node, encouraging the model to focus on 
more informative node pairs. Below, we summarize the major contributions of this work.

\begin{itemize}
    \item First, we design a joint video graph that models neighbor relations in space and similarity relations in time  for visual correspondence learning.
    \item Second, we formulate the contrastive learning as a random but attentive walk on the graph to learn discriminative representations from seeing both temporal and neighboring views of instances.
    \item Third, our method outperforms state-of-the-art self-supervised approaches on a variety of visual tasks, \eg, object, part propagation, and pose tracking. It also surpasses some task-specific fully supervised algorithms. 
\end{itemize}

\section{Related Works}

\paragraph{Self-supervised Representation Learning.}
Learning visual representations from unlabeled images or videos has been widely explored in many \textit{pretext} tasks, including future prediction~\cite{srivastava2015unsupervised, lotter2016deep}, frame sorting~\cite{lee2017unsupervised, misra2016shuffle}, motion estimation~\cite{agrawal2015learning, tung2017self}, and audio analysis~\cite{owens2018audio, korbar2018cooperative}. These methods learn good feature representations that can generalize well to multiple tasks by further fine-tuning on a small set of labeled samples. The key idea in them is to utilize the inherent information inside images or videos as the supervisory signals. For example, corresponding pairs can be constructed by augmentation of the same instance~\cite{wu2018unsupervised, bojanowski2017unsupervised}. However, manually augmenting still images may not always be in correct correspondence.
Recent works in \textit{contrastive learning}~\cite{oord2018representation, chen2020simple, he2020momentum, gordon2020watching} explore the supervisory signals for similarity learning by choosing pairs that are close in space~\cite{he2020momentum, chen2020simple, bachman2019learning} or time~\cite{gordon2020watching, oord2018representation, sermanet2018time}. In contrast, we implicitly determine which pairs to be closer by their neighbor relations in space and similarity relations in time.

\begin{figure*}
	\centering
	\includegraphics[width=0.9\textwidth]{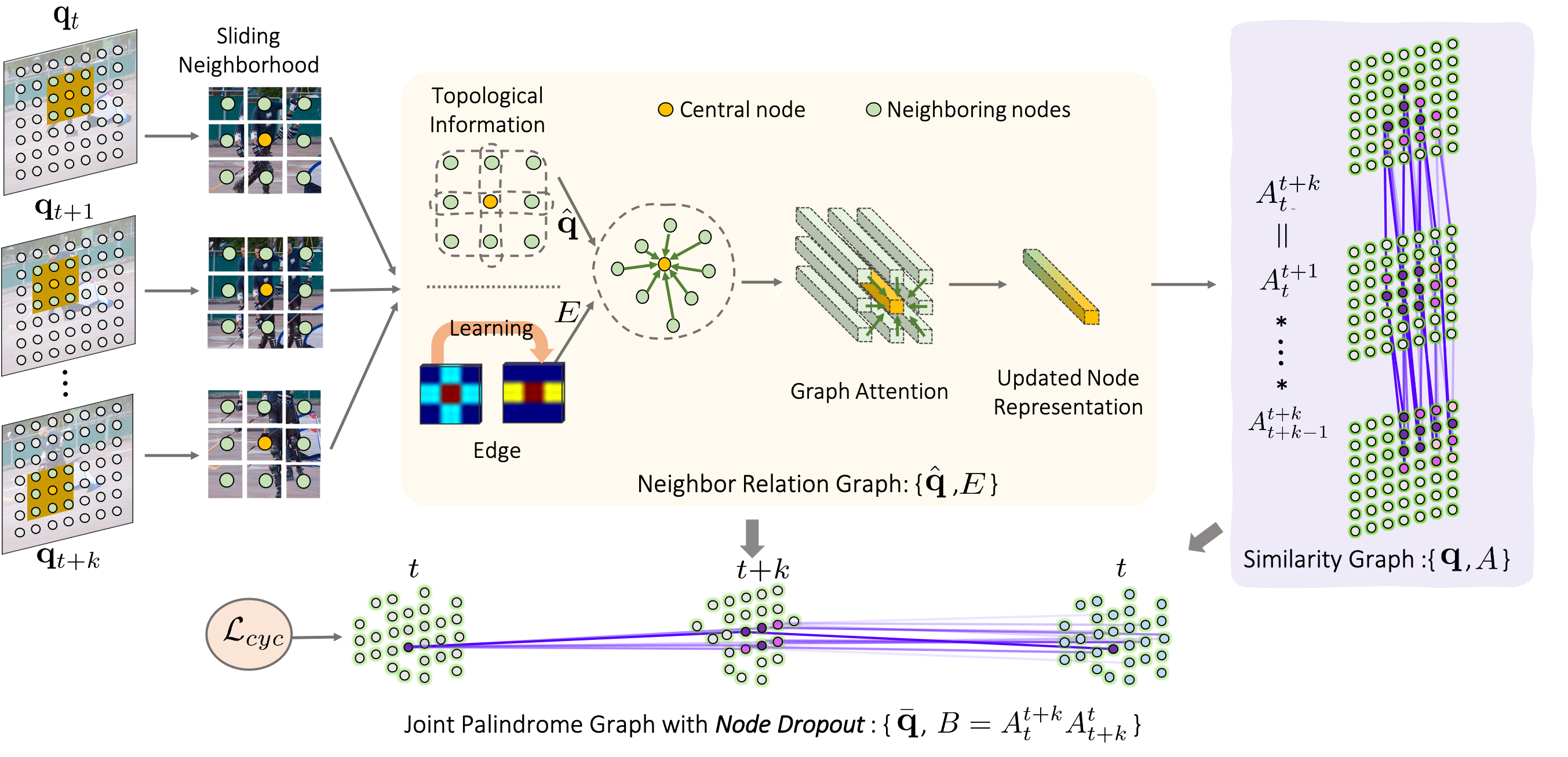}
\caption{Schematic illustration of our joint space-time graph for correspondence learning. Specifically, we have two sub-graphs that associate grid patches (nodes) with different relations. (i) Neighbor Relation Graph $\{\hat{\mathbf{q}}, E\}$: it connects a central node to its neighbors $\hat{\mathbf{q}}$ with edge $E$ initialized with topological prior, by which the neighboring embeddings can be aggregated to the center in a learnable manner. (ii) Similarity Graph $\{\mathbf{q}, A\}$: it links inter-frame nodes $\mathbf{q}$ with pair-wise similarity affinities $A$ (in updated representation space) to form a multi-step association on a long-range sequence. Furthermore, we employ the node dropout technique and transfer sequence as palindrome to upgrade graph to $\{\bar{\mathbf{q}}, B\}$, in which we perform a random but attentive walk to find the correspondence  based on contrastive learning.} 
\label{fig:method}
\end{figure*}

\vspace{-2mm}
\paragraph{Self-supervised Correspondence Learning.} Recent approaches focus on learning correspondence from unlabeled videos in a self-supervised manner. 
The key idea of TimeCycle~\cite{wang2019learning} is to train a deterministic patch tracker to find the correspondence of a query patch by tracking forward and backward in the video. Likewise,  UVC~\cite{li2019joint} and ContrastCorr~\cite{Wang_2021_Contrastive} adopt the patch tracker to obtain  object-level correspondences. But they also explore fine-grained correspondences by learning a pixel-wise affinity with colorization. The difference is that Wang \etal~\cite{Wang_2021_Contrastive} combines the intra-video transformation~\cite{li2019joint} with inter-video transformation to form contrastive pairs. Besides, CorrFlow~\cite{lai2019self} and MAST~\cite{lai2020mast} use feature maps with higher resolution ($2\times$) than others and yield impressive results. 
Recently, Jabri \etal~\cite{jabri2020walk} formulate the correspondence as a contrastive random walk, allowing associations between patches that may have significant differences in appearance. Despite the success of these methods, many of them still struggle with the overwhelming noisy or negative samples when performing query-target matching. Our approach tackles this issue by introducing neighboring views to the matching pairs for contrastive learning, which allows us to lean implicit associations between central representation and its neighbor.

\vspace{-1mm}
\paragraph{Video Graphs.}
Representing video as graphs can usually capture the spatial-temporal relationships in videos~\cite{tsai2019video, wang2018videos, qian2019video, jabri2020walk}. The key of video graph is to form the image patches as nodes and link them with edges.
One popular direction is to model the object-object interactions by connecting objects which overlap in space or close in time. They have been widely applied to video classification~\cite{wang2018videos}, detection~\cite{qian2019video}, or visual relationship reasoning~\cite{tsai2019video}, by combining with Conditional Random Fields (CRF)~\cite{lafferty2001conditional} or Graph Convolutional Networks (GCN)~\cite{kipf2016semi}. Recently, some works start to model how the states of the same object change in time by connecting inter-frame nodes that have similar appearance or semantically related~\cite{jabri2020walk, wang2018videos}. Leveraging the similarity relations, the task of representation learning can be induced by propagating the node identity in a graph. To better learn instance discrimination with cross-attention between nodes, we go one step further by modelling the neighbor relations of intra-frame nodes, which allow us to learn latent associations between central node and its neighbors across space and time.

\section{Methods}

We propose to represent the video as a joint space-time graph for learning temporal correspondence.
As shown in Figure~\ref{fig:method}, nodes are frame patches sampled in a grid, and edges contain two type of connections: neighbor relations between intra-frame nodes, and visual similarity between inter-frame nodes. Based on these two relations, the big graph can be decomposed into two sub-graphs, including neighbor relation graph and similarity graph, aiming at capturing the \textit{broader} and \textit{longer} views, respectively.
Next, we perform reasoning on the graph to find latent correspondences for contrastive learning.
Our learning process can be interpreted as a random but attentive walk on the graph by automatically dropping out some ``common-fate'' nodes.
The construction details of  two sub-graphs and learning procedures are successively described in this section.

\subsection{Neighbor Relation Graph}
Given a  video sequence $\mathbf{I}$, we denote a set of nodes $\mathbf{q}_t$ that correspond to $N$ overlapped patches sampled in a grid from frame $\mathbf{I}_t$. In general, each node in $\mathbf{q}_t$ will be mapped to a $l_2$-normalized $d$-dimensional embedding using an encoder $\phi$, where $d$ is the channel number. The embedding captures a very local view of an object sometimes, although necessary for learning tiny structures, it easily induces ambiguity into the  query-target matching. 
Intuitively, the neighboring nodes provide the central node with a broader view of an object or interactions with other objects, which are beneficial to find its temporal correspondence. 
Motivated by this, we build a neighbor relation graph that reinforces node embeddings by relating the information from the neighbors guided by the general topology.
The edge $E$ is solely established between node $i$ and its neighbors rather than all other nodes~\cite{ma2020copulagnn}.
The sum of all edge values connected to node $i$ is normalized to be 1 by a softmax function. If node $i$ and $j$ are spatially closer or more correlated in semantics, then $E_{ij}$ should be higher.
We denote the neighbor relation graph as $G_r = \{\hat{\mathbf{q}}, E\}$ where $\hat{\mathbf{q}}$ is the set of $n$ neighbors.
\begin{figure}
	\centering
	\includegraphics[width=85mm]{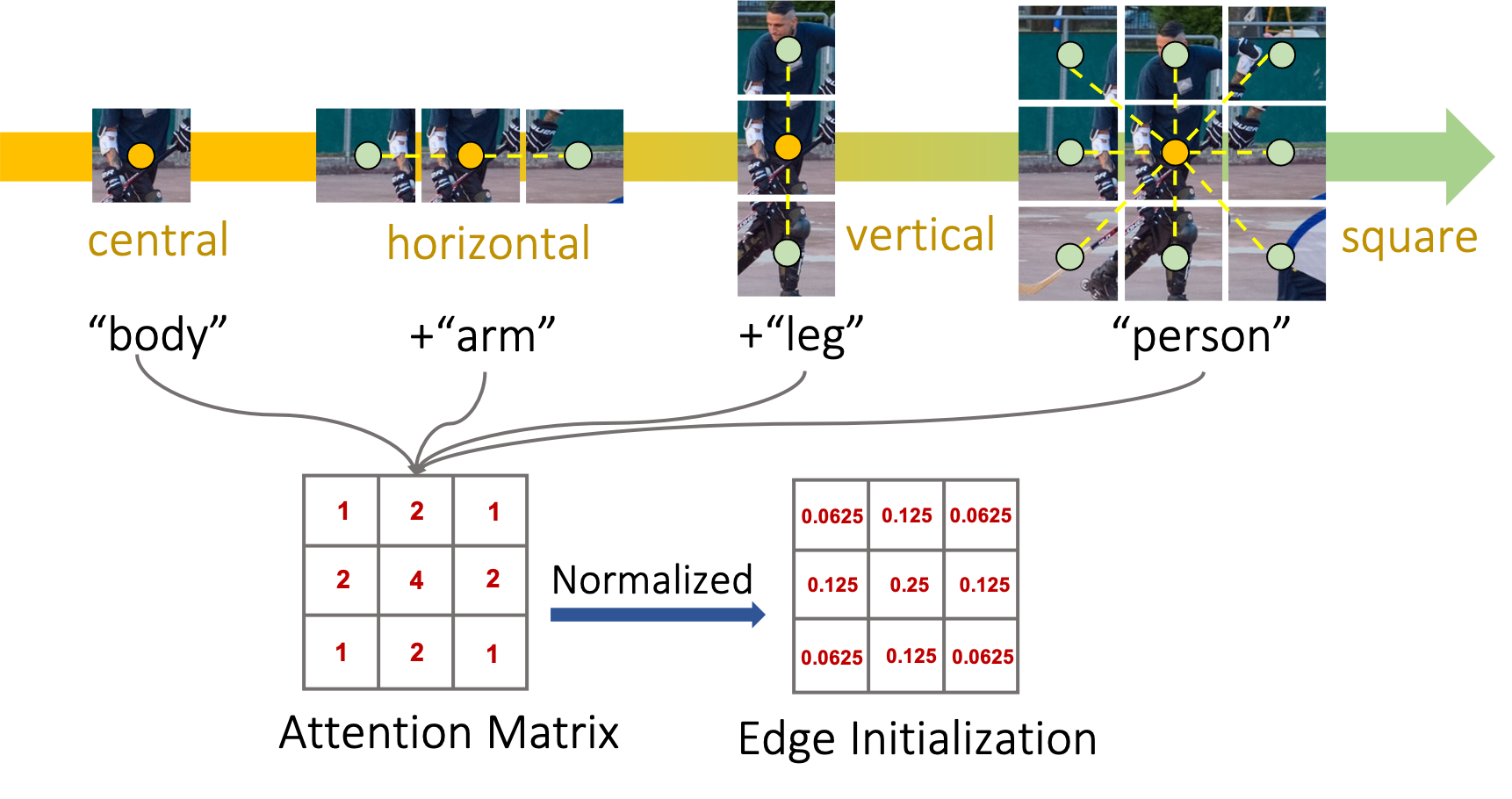}
\caption{\textbf{Encoding topological information} in the neighbor relation graph by edge initialization. Three structures (\ie, horizontal, vertical, square) associate the central node to broader representations variously.  We adopt the normalized attention matrix as the initial value of edges to encode the topological prior. }
\label{fig:init}
\end{figure}

\noindent \textbf{Sliding neighborhood.} We construct $G_r$ by considering a friendly neighborhood~\cite{hu2019local, kim2021find}  as a small grid (\eg, a $3 \times 3$ grid yielding 9 neighbors), and model the neighbor relations for the central node. A larger neighborhood is not necessary because farther nodes are more likely to induce noise (see performance degradation in Figure~\ref{fig:ev} (a)). 
For different nodes in $\mathbf{q}_t$, we consider them as the center and determine corresponding neighborhoods in a sliding window manner. This lead to a shared  edge $E \in \mathbb{R}^{N\times n}$ across $N$ nodes in $\mathbf{q}_t$, \ie, $E_{1j} = E_{2j} = ... = E_{Nj}$, where $j\in \{1,2,...,n\}$.

\noindent \textbf{Encoding topological prior.} To model the neighbor relation, we first initialize $E$ by explicitly encoding the topological information. 
In a neighborhood (see Figure~\ref{fig:init}), we generally have three types of topology with regard to the central node based on the spatial proximity, including vertical, horizontal, and square structures. Each of them may composite central element into a higher-level entity in the feature space with neighboring views. For example, a ``body center'' can be extended with semantics of ``arm'', ``legs'', or ``person'' using horizontal, vertical, or square topology. We do have other cases that object-object interactions are captured by the neighbors, which also provide an crucial cue in modelling neighbor relations. 
To this regard, we generate a normalized attention matrix based on the number of node occurrences in different topology, and employ the matrix to initialize edge $E$. The topological prior can therefore be encoded to guide the following learning of $E$.

\noindent \textbf{Graph attention.} Advanced by the topological information, the graph captures the \textit{relational importance} of neighboring nodes, in other words, the degree of importance of each of the neighbors contributing to the central node. 
We then explore a graph attention mechanism to augment the representations of the central node $i$ by aggregating messages from its neighboring nodes:
\begin{equation}
   f(\mathbf{q}_t^i) = \sum_{j=0}^{n} \texttt{softmax}(E_{ij})\cdot\phi(\mathbf{q}_t^{j}),
   \label{eq:0}
\end{equation}
where $\mathbf{q}_t^j$ is the $j$-th node in $\mathbf{q}_t$ and $f(\mathbf{q}_t^i)$ is the updated embedding of node $\mathbf{q}_t^i$, which provides weighted neighboring semantics, while preserving the original feature patterns. 
Different from the channel-level feature aggregation with GCN~\cite{wang2018videos, kipf2016semi}, our graph attention performs a node-level feature simulation that treats each node embedding as a whole.
We show this mechanism benefits the contrastive learning in Section~\ref{sec:learning}. More importantly, $E$ is learnable via back propagation, by which a more general neighbor relation can be modeled  during training.

\subsection{Similarity Graph}
After considering the intra-frame node relations, we link the visually corresponding nodes in the adjacent frames by a similarity-based affinity. One general option for the pairwise similarity function is a dot-production between two feature embeddings: $F (\phi(q_1), \phi(q_2)) = \phi(q_1)^{\top} \phi(q_2) $. Following recent similarity learning methods~\cite{ jabri2020walk, li2019joint, Wang_2021_Contrastive}, we employ a row-wise softmax function to the similarity function with temperature $\tau$ to obtain a non-negative affinity matrix between inter-frame node embeddings updated by $G_r$:
\begin{equation}
    A_t^{t+1}(i,u)= \frac{\text{exp}(F(f(\mathbf{q}_t^i),~f(\mathbf{q}_{t+1}^u))/\tau)}{\sum_{l=1}^N\text{exp}(F(f(\mathbf{q}_t^i),~f(\mathbf{q}_{t+1}^l))/\tau)}.
    \label{eq:1}
\end{equation}

The affinity in Equation~\eqref{eq:1} places weights on all possible edges between nodes at $t$ and $t+1$, with higher possibility indicating that the paired patches are more similar. Given such connections, we can already construct a simple similarity graph  between two adjacent frames in the video. However, the short temporal dynamics within two frames provides very limited views of objects.
We therefore link all the inter-frame nodes across a video with a length of $T$, and formulate the graph path as a Markov chain of edges, following the idea in~\cite{jabri2020walk}:
\begin{equation}
A_t^{t+T} = \prod_{i=0}^{\mathrm{T}-1} A_{t+i}^{t+i+1}.
\label{eq:Gs}
\end{equation}
Here, we can denote the similarity graph as $G_s = \{ \mathbf{q}, A\}$ where $\mathbf{q}$ represents all inter-frame nodes in a video and $A$ is a chain of edges described in Equation~\eqref{eq:Gs}.

\subsection{Attentive Walk on Joint Space-Time Graph}
\label{sec:learning}
Our goal is to learn temporal correspondence in the joint space-time graph without human annotations. Akin to prior arts that explore cycle-consistency in time~\cite{wang2019learning, jabri2020walk}, we adopt the \textit{palindrome}  sequence in the form of \{$I_t,...,I_{t+T},...,I_t$\} for training, where the target of the query node should be its original position. 
Following the idea of~\cite{jabri2020walk}, we build our cycle-consistent loss as:
\begin{equation}
    \mathcal{L}_{cyc}(G | G = \{ \mathbf{q}, B\}) =\mathcal{L}_{CE} (B, I),
\end{equation}
where $G$ is a  joint palindrome graph with edges $B = A_t^{t+T}A_{t+T}^t$, and $I$ is the target position generated according to the location of the first frame node, \eg, the ground truth of the $i$-th node is $i$.

\begin{figure}
	\centering
	\includegraphics[width=85mm]{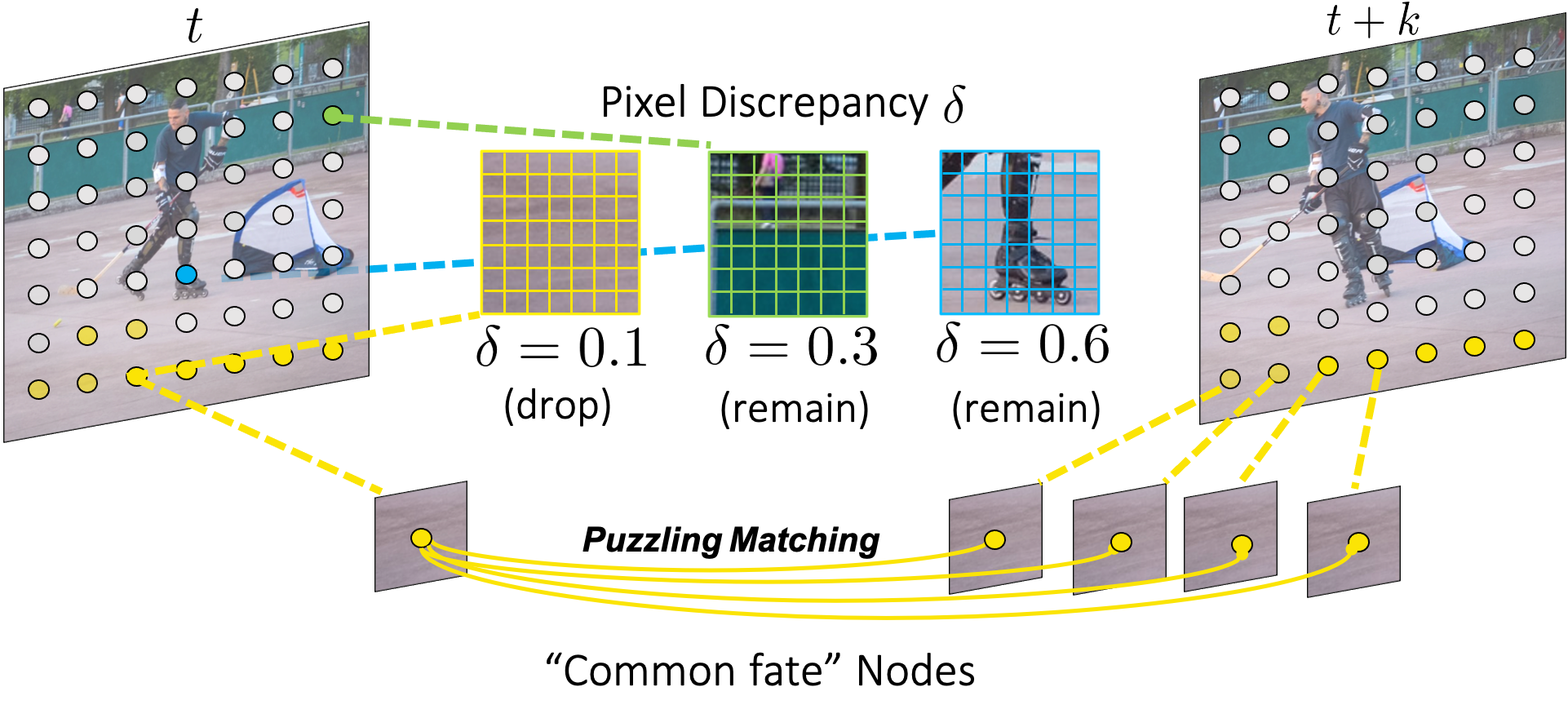}
\caption{\textbf{Node dropout} to avoid ``common-fate'' nodes that puzzle the correspondence learning. We introduce a thresholding measurement based on the pixel discrepancy $\delta$. }
\label{fig:drop}
\end{figure}

\noindent \textbf{Node dropout.}
Compared with \textit{things} --- countable objects such as people and animals, one problem of learning correspondence in graphs is matching \textit{stuff} --- large regions of similar textures or materials, such as sky and land (Figure~\ref{fig:drop}).
The ``common-fate''~\cite{wertheimer1938laws} nodes in the stuff have strong affinity to all other nodes in the related segment of neighboring frames, making it hard and somewhat impossible to walk back to the original position during training. 
To address this issue, we propose a node dropout strategy based on the pixel discrepancy of a node, given that the ``common-fate'' nodes always contain very similar context.
Specifically, we start by retrieving the pixel embeddings $p \in \mathbb{R}^{d\times hw}$ for each node using the encoder $\phi$, where $hw$ is the downsampled spatial size. 
Next, we calculate the self-similarity  among pixel embeddings via a dot-production: $S =p^{\top}p$. We define the pixel discrepancy of a node as:
\begin{equation}
   \delta = 1- \frac{1}{(hw)(hw)}\sum_{i=0}^{hw}\sum_{j=0}^{hw}S_{ij},
\end{equation}
where we convert the self-similarity to the reverse side, so that the higher value of $\delta$ denotes higher discrepancy among pixels. We then set a threshold of $\delta$ to drop those uninformative nodes whose pixel discrepancy is lower than it. The proposed strategy is superior to the random dropout technique in~\cite{jabri2020walk} by exclusively tackling ``common-fate'' nodes.
Our final training objective is:
\begin{equation}
    \mathcal{L}_{cyc}(\bar{G} | \bar{G} = \{ \bar{\mathbf{q}}, B^k\}) = \sum_{k=1}^\mathrm{T}\mathcal{L}_{CE} (B^k, I),
    \label{eq:3}
\end{equation}
where $\bar{\mathbf{q}}$ is the remained nodes after node dropout, and $B^k = A_t^{t+k}A_{t+k}^t$. We optimize all  sub-cycles in the graph with the clip length $k$ varying from 1 to T.  
In this regard, we are allowed to perform a random but attentive walk on the graph, forcing the model to discriminate informative node pairs by the hedging of ambiguous matching.

\noindent \textbf{Contrastive learning with \textit{extra} positive pairs.}
We can consider our model as a chain of contrastive learning problem, 
which is guided by a ``one-hop'' cycle-consistency constraint. Basically, a strong edge creates a one-to-one alignment, \ie, one positive pair. However, after aggregating  neighboring information via Equation~\eqref{eq:0}, we can interpret one edge as a \textit{latent} many-to-many alignment in the feature space, involving \textit{extra} positive pairs for contrastive learning apart from center-center pairs.
For example, center-neighbor pairs --- a node at timestep $t$ and one of its corresponding neighbors at timestep $t+1$, or even neighbor-neighbor pairs.
Taking the hand-arm pair as an example, since they are physically connected as neighbors in most cases, when learning the correspondence of ``hand'', our model may push its embedding closer to that of ``arm'' as well.
Recall that we learn $E$ in Equation~\eqref{eq:0} via the objective~\eqref{eq:3}. The learned $E$ encourages the model to find more reliable center-center or center-neighbor pairs for contrastive learning, which  generate better node representations in return for $E$ to model the general neighbor relations. We believe this is the main reason that our model learns more discriminative representations of instances.

\section{Experiments}
We extensively evaluate our learned representations on various visual correspondence tasks: video object propagation, human part propagation, and pose keypoint tracking. 
We first conduct the comparison with the state-of-the-art self-supervised algorithms for visual correspondence learning, including TimeCycle~\cite{wang2019learning},CorrFlow~\cite{lai2019self}, MAST~\cite{lai2020mast}, UVC~\cite{li2019joint}, ContrastCorr~\cite{Wang_2021_Contrastive}, and VideoWalk~\cite{jabri2020walk}.
We then compare our model with pre-trained features from representation learning methods, 
including MoCo~\cite{he2020momentum}, a self-supervised contrastive learning method based on images; VINCE~\cite{gordon2020watching}, an extension of MoCo to videos; 
ImageNet~\cite{he2016deep}, a strong supervised method where the model is pre-trained on ImageNet. All above methods use ResNet-18~\cite{he2016deep} as the backbone. Besides, 
we also compare with some fully supervised algorithms designed for the specific tasks.
At last, we provide in-depth ablation studies.

\subsection{Implementation}
\begin{figure*}
	\centering
	\includegraphics[width=170mm]{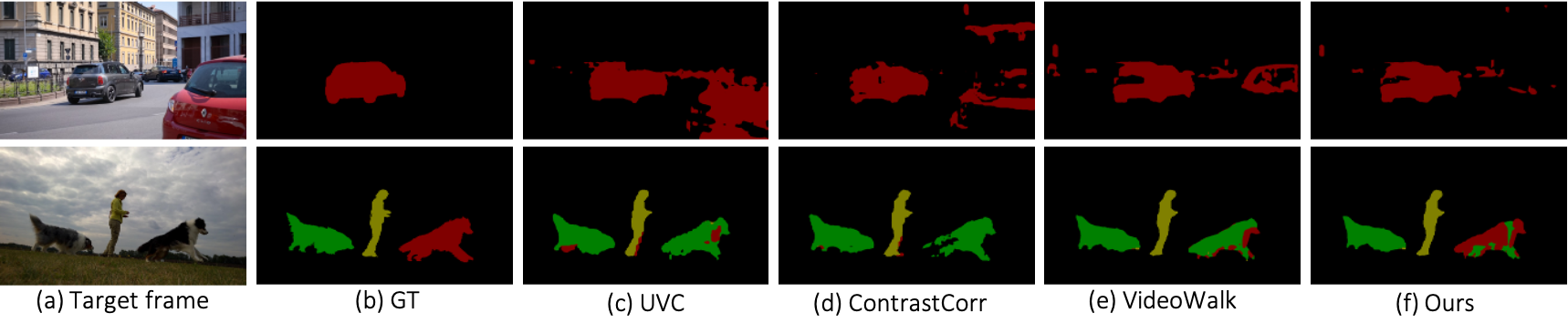}
\caption{Qualitative comparisons with other self-supervised methods on DAVIS 2017 dataset. (a) Target frame. (b) Ground-truth of target frame. (c) Results of UVC~\cite{li2019joint}. (d) Results of ContrastCorr~\cite{Wang_2021_Contrastive}. (e) Results of VideoWalk~\cite{jabri2020walk}. (f) Our results.}
\label{fig:compare}
\end{figure*}

\begin{table*}[!t]
\caption{\textbf{Video object propagation} results on DAVIS 2017 dataset. We show results of state-of-the-art self-supervised methods and some supervised approaches in comparison of our method. \textit{Train Data} indicates dataset(s) used for pre-training, including:
I = ImageNet~\cite{he2016deep}, K = Kinetics400~\cite{carreira2017quo}, C = CoCo~\cite{lin2014microsoft}, D = DAVIS 2017~\cite{pont20172017}, P = PASCAL-VOC~\cite{everingham2015pascal}, Y = YouTube-VOS~\cite{xu2018youtube}, O = OxUvA~\cite{valmadre2018long}, V = VLOG~\cite{fouhey2018lifestyle}, T = TrackingNet~\cite{muller2018trackingnet}. \textit{Resolution} indicates whether feature map for correspondence matching is of a higher ($2\times$) resolution.}
	\label{tab:result}
	\centering
\resizebox{150mm}{!}{
\begin{tabular}{lccccccccc}
\hline
Method  & Supervised  & Backbone   & Train Data  & Resolution & $\mathcal{J}$\&$\mathcal{F}_m$    &$\mathcal{J}_m$      & $\mathcal{J}_r$     &  $\mathcal{F}_m$    & $\mathcal{F}_r$     \\ \hline

MoCo~\cite{he2020momentum}  &   & \multirow{2}{*}{ResNet-18}   & I    & \multirow{2}{*}{1$\times$ }    & 60.8 & 58.6 & 68.7 & 63.1 & 72.7 \\
VINCE~\cite{gordon2020watching}  &  &    & K        &         & 60.4 & 57.9 & 66.2 & 62.8 & 71.5 \\ \hdashline
CorrFlow~\cite{lai2019self} &  &\multirow{3}{*}{ResNet-18}  & O       & \multirow{3}{*}{2$\times$ }          & 50.3 & 48.4 & 53.2 & 52.2 & 56.0 \\
MAST~\cite{lai2020mast}   &    &   & O       &          & 63.7 & 61.2 & 73.2 & 66.3 & 78.3 \\
MAST~\cite{lai2020mast}   &   &   & Y      &         & 65.5 & 63.3 & 73.2 & 67.6 & 77.7 \\ \hdashline
TimeCycle~\cite{wang2019learning} & &\multirow{4}{*}{ResNet-18}  & V        & \multirow{4}{*}{1$\times$ }         & 48.7 & 46.4 & 50.0 & 50.0 & 48.0 \\
UVC~\cite{li2019joint}    &   &   & K        &         & 60.9 & 59.3 & 68.8 & 62.7 & 70.9 \\
ContrastCorr~\cite{Wang_2021_Contrastive}& & & T &         & 63.0   & 60.5 & 70.6    & 65.5 & 73.0    \\
VideoWalk~\cite{jabri2020walk} & &  & K       &         & 67.6 & 64.8 & 76.1 & 70.2 & 82.1 \\ \hdashline
Ours    &  &  ResNet-18 & K        & 1$\times$         &\textbf{68.7} & \textbf{65.8} & \textbf{77.7}  &\textbf{71.6} & \textbf{84.3}   \\ \hline
ImageNet~\cite{he2016deep} &$\checkmark$  &ResNet-18  & I    & 1$\times$         & 62.9 & 60.6 & 69.9 & 65.2 & 73.8 \\
SiamMask~\cite{wang2019fast} &$\checkmark$& ResNet-50 & I/C/Y    & \multirow{5}{*}{ }        & 56.4  & 54.3 & 62.8 & 58.5 & 67.5 \\
OSVOS~\cite{caelles2017one}&$\checkmark$ & VGG-16 & I/D   &         & 60.3  & 56.6 & 63.8 & 63.9 & 73.8 \\
OnAVOS~\cite{voigtlaender2017online}&$\checkmark$ & ResNet-38 & I/C/P/D   &        & 65.4 & 61.6 & 67.4 & 69.1 & 75.4 \\
OSVOS-S~\cite{maninis2018video} &$\checkmark$& VGG-16 & I/P/D   &         & 68.0 & 64.7 & 74.2 & 71.3 & 80.7 \\\hline
\end{tabular}
}
\vspace{-1mm}
\end{table*}

\noindent \textbf{Encoder.} For fair comparisons, we also adopt ResNet-18~\cite{he2016deep} as the encoder $\phi$ by reducing the stride of last two residual blocks (\texttt{res3} and \texttt{res4}) to 1.
We add a linear projection after average pooling to generate a 128-dimensional node embedding. 
Pixel-wise embeddings of each node are created by the same projection without average pooling for node dropout.

\noindent \textbf{Training.} We train $\phi$ using the unlabeled videos from Kinetics400~\cite{carreira2017quo} dataset with the Adam optimizer. We set the temperature $\tau$ as 0.05 in Equation~\ref{eq:1}. 
Akin to~\cite{jabri2020walk}, for each $256\times 256$ frame, we sample $64\times 64$ patches in a $7\times 7$ grid, resulting in 49 nodes per frame.
Without extra specification, our sliding neighborhood is a $3\times3$ grid, involving 9 neighboring nodes, and the length of training sequences is 10. To see sufficient samples, we first train the model without node dropout for 5 epochs using a learning rate of $1\times 10^{-4}$. Next, we set $\delta=0.2$ for node dropout and train the model for another 15 epochs with a learning rate of $1\times 10^{-5}$.  All experiments are conducted on 4 NVIDIA Titan Xp GPUs.

\noindent \textbf{Inference.} All evaluation tasks can be considered as video label propagation, which is to predict the labels of each pixels in the target frame given only the labels in the first frame (\ie, the source).
For fair comparisons, we use the same label propagation strategy and testing protocols as~\cite{jabri2020walk} for all tasks. In brief, labels $L_t$ are propagated as $L_t = K_t^s L_s$, where $L_s$ is the source labels and $K_t^s$ is the top-$k$ transitions between source and target frames ($k$ is 10 for all tasks). To provide temporal context, the last $m$ frames are also used for propagation ($m$ is 20, 4, and 7 for DAVIS, VIP and JHMDB tasks respectively). To avoid noise from pixels that are far away in space, the query pixels are restricted by a \textit{local} attention mask with a radius $r$ ($r$ is 5 for JHMDB, and 12 for all other tasks).
We use the output of \texttt{res3} as feature representations to calculate affinities for label propagation, and $\tau$ is set as 0.05 for consistency with the training.

\subsection{Video Object Propagation on DAVIS 2017}
We evaluate our model on a popular benchmark of semi-supervised video object segmentation, \ie, DAVIS 2017~\cite{pont20172017}, which provides the semantic mask of multiple objects in the first frame. For fair comparisons with prior works~\cite{jabri2020walk, Wang_2021_Contrastive, li2019joint, wang2019learning}, we test the model on images with the resolution of 480p.
We report the mean (m) and recall (r) of Jaccard index $\mathcal{J}$ (IoU) and contour alignment $\mathcal{F}$, detailed in Table~\ref{tab:result}. 
Figure~\ref{fig:compare} and Figure~\ref{fig:results} (a) show the propagated object masks. 
Specifically, our approach attains improvements over MoCo~\cite{he2020momentum} and VINCE~\cite{gordon2020watching}, indicating that it is better to choose temporal views for contrastive learning in videos rather than data augmentation of a single frame.
Our method also performs favourably against the self-supervised methods from unlabeled video, and even without relying on a higher-resolution feature map used in CorrFlow and MAST~\cite{lai2019self,lai2020mast} or other modules such as the patch localizer designed in UVC and ContrastCorr~\cite{li2019joint, Wang_2021_Contrastive}.
Our method achieves consistent improvements over the state-of-the-art method VideoWalk~\cite{jabri2020walk} across all the evaluation metrics.
Surprisingly, our method can outperform many supervised methods~\cite{wang2019fast, caelles2017one, voigtlaender2017online, maninis2018video} with specific architectures for video object segmentation. 

Furthermore, we show that by modelling neighbor relations during training, our model exhibits  superior discrimination capability for instance-level separation. As seen in Figure~\ref{fig:compare}, both UVC~\cite{li2019joint} and ContrastCorr~\cite{Wang_2021_Contrastive} fail to discriminate similar instances by learning pixel-level correspondence. Despite seeing more hard negative samples during training, the walker in~\cite{jabri2020walk} still gets confused to instances that are similar in color. In contrast, our model can tell the difference of small parts between similar instances, \eg, the dog tails or the car windows, by inducing neighboring views for contrastive learning.

\begin{table}[!t]
\caption{\textbf{Part segmentation and Pose tracking} results on VIP and JHMDB datasets, respectively. We compare our model  with  self-supervised and strong supervised methods. \textit{Sup} indicates it is a supervised method or not. }
	\label{tab:result1}
	\centering
	\resizebox{85mm}{!}{
\begin{tabular}{lcccc}
\hline
\multirow{2}{*}{Method} & \multirow{2}{*}{Sup} & \multicolumn{2}{c}{Pose} & \multicolumn{1}{c}{Part} \\ \cmidrule(r){3-5} 
                 &       & PCK@0.1     & PCK@0.2    & mIoU       \\\hline
TimeCycle~\cite{wang2019learning}  &  & 57.3        & 78.1       & 28.9             \\
UVC~\cite{li2019joint}  &   & 58.6        & 79.6       & 34.1            \\
ContrastCorr~\cite{Wang_2021_Contrastive} &  & 61.1        & 80.8       & 37.4             \\ 
VideoWalk~\cite{jabri2020walk} & & 59.3        & 84.9       & 38.6               \\
Ours               &     &  \textbf{61.4}   & \textbf{85.3}    &  \textbf{40.2}       \\\hline
ImageNet~\cite{he2016deep} &$\checkmark$ & 53.8        & 74.6       & 31.9             \\
ATEN~\cite{zhou2018adaptive}  & $\checkmark$  & -           & -          & 37.9             \\
Thin-Slicing Net~\cite{song2017thin}  & $\checkmark$ & 68.7        & 92.1       & -                 \\ \hline
\end{tabular}
}
\vspace{-1mm}
\end{table}

\subsection{Human Part Propagation on VIP}
We evaluate our method on part segmentation task on the Video Instance Parsing (VIP) benchmark~\cite{zhou2018adaptive}, which
involves propagating  20 parts of human (\eg, arms and leg), requiring more precise matching than DAVIS. 
We use the same settings as Jabri \etal~\cite{jabri2020walk}, and resize the video frames to $560\times 560$. For the semantic part propagation task, we evaluate performance via the mean IoU metric.
As seen in Table~\ref{tab:result1}, our model outperforms existing self-supervised methods, \eg, by 1.6\% and  2.8\% mIoU, compared to VideoWalk~\cite{jabri2020walk} and ContrastiveCorr~\cite{Wang_2021_Contrastive}, respectively. We also surpass the fully supervised method ATEN~\cite{zhou2018adaptive} that is specifically designed for this dataset using training labels.
Figure~\ref{fig:results} (b) shows samples of semantic part propagation results. Interestingly, our model correctly propagates each part mask onto similar instances (dancers in the first example) no matter when they are close or far from the camera.

\begin{figure}
	\centering
	\includegraphics[width=85mm]{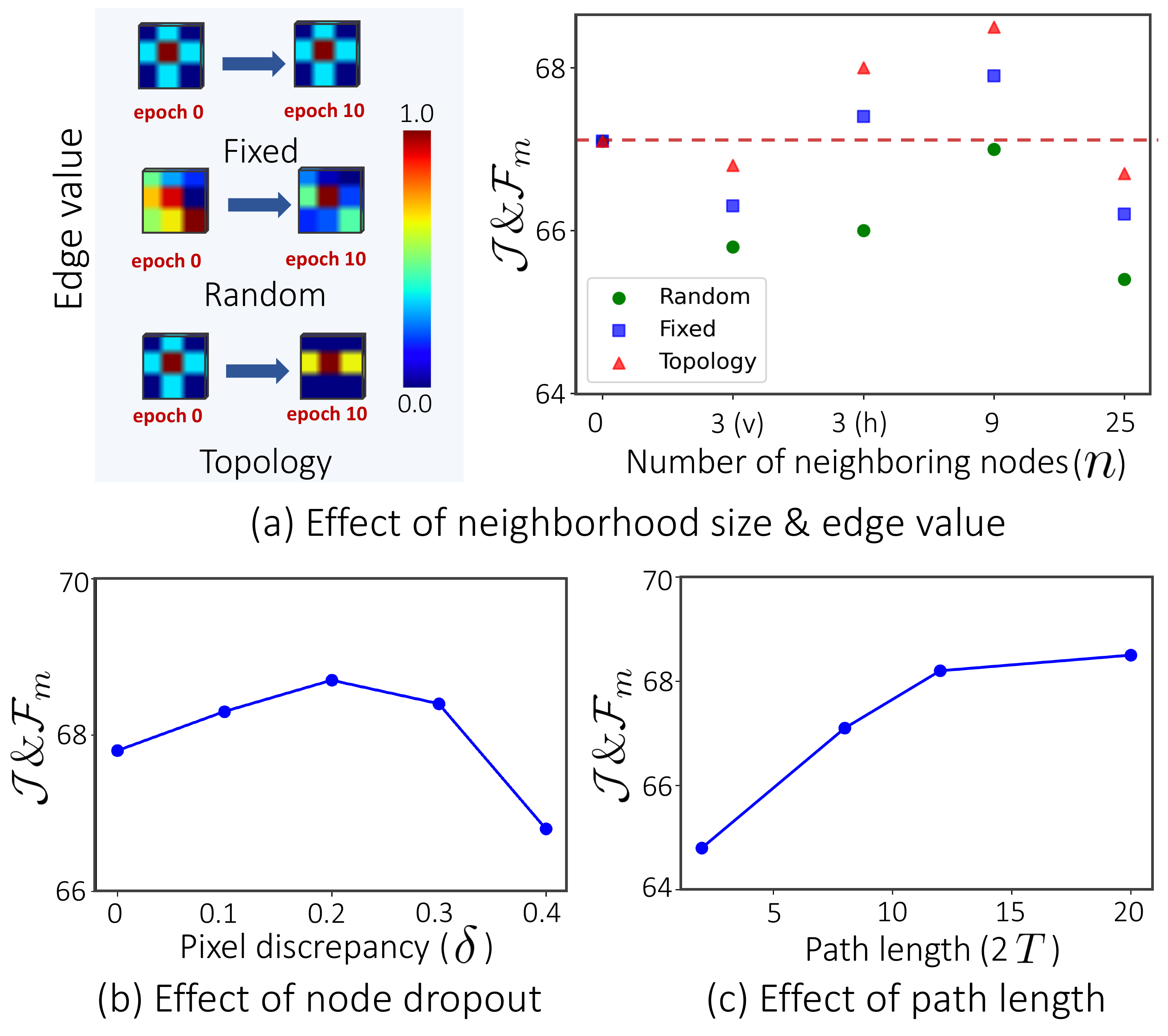}
\caption{Ablation studies of our method on DAVIS 2017 benchmark. (a) Effect of neighborhood size and edge value. (b) Effect of node dropout. (c) Effect of training path length.}
\label{fig:ev}
\vspace{-2mm}
\end{figure}

\subsection{Pose Keypoint Tracking on JHMDB}
\begin{figure*}
	\centering
	\includegraphics[width=0.99\textwidth]{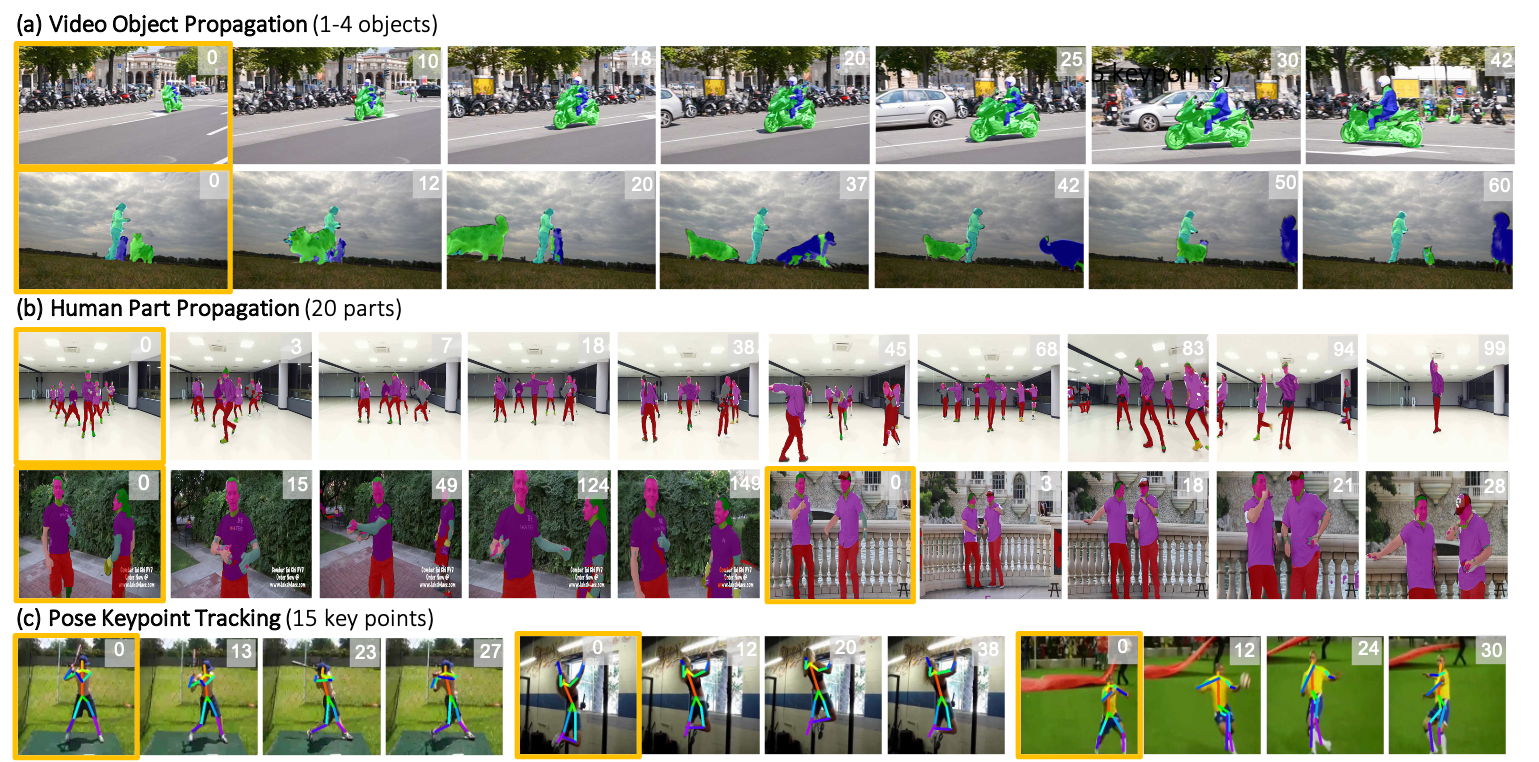}
\caption{Propagation results of our model. (a) Video object propagation on DAVIS 2017~\cite{pont20172017} dataset.  (b) Human part propagation on VIP~\cite{zhou2018adaptive} dataset. (c) Pose keypoint tracking on JHMDB~\cite{jhuang2013towards} dataset. The first frame is highlighted with a yellow outline with its label being provided. Without fine-tuning, our model achieves promising long-range label propagation on three visual tasks.  }
\label{fig:results}
\end{figure*}

We consider the pose tracking task on JHMDB benchmark~\cite{jhuang2013towards}, which involves 15 keypoints. Follow the evaluation protocol of~\cite{li2019joint,jabri2020walk}, we test the model on $320\times 320$px images. We adopt the possibility of correct keypoints~\cite{song2017thin} (PCK) as the evaluation metric, which measures the percentage of keypoints close to ground-truth under different thresholds. We show quantitative evaluations against others in  Table~\ref{tab:result1}, and qualitative results in Figure~\ref{fig:results} (c). Our model achieves consistent improvements over existing self-supervised approaches on this challenging task that requires precise fine-grained matching.  
Notably, our model achieves even 10.7\% better in PCK@0.2 than the ImageNet~\cite{he2016deep} baseline trained with classification labels.


\subsection{Analytical Ablation Studies on DAVIS 2017}

\noindent \textbf{Neighborhood size.} We investigate how necessary to relate more neighboring nodes for a broader view, by constructing neighbor relation graph with dimension of $3\times1$, $1\times3$, $3\times3$, $5\times5$ --- resulting in $n=$ 3, 3, 9, 25 nodes, respectively.
In Figure~\ref{fig:ev} (a), we find that  9 neighboring nodes can peak the performance on DAVIS. Further increasing the neighborhood size ($n=25$) is likely to induce noisy cues from farther nodes, resulting in even worse results than the baseline that does not consider neighbor relations.
Interestingly, the interactions with horizontally-connected nodes are more beneficial to learn discriminative representations than the vertical ones (see 3 (h) \vs 3 (v)). 

\noindent \textbf{Variants of edge $E$.} We also explore three variants of edge $E$ in the neighbor relation graph in Figure~\ref{fig:ev} (a): (i) \textit{Fixed}: fixed edge with topological information; (ii) \textit{Random}: learnable edge with random initialization; (iii) \textit{Topology}: learnable edge with topological initialization. In conclusion, encoding topology is essential for modelling neighbor relation of nodes. Further learning edge $E$ at training time yields better results. We attribute this success by more general neighbor relations gained in learning procedures.

\noindent \textbf{Node dropout.}
We evaluate the effect of node dropout by training our model with ranging values of $\delta \in$ [0, 0.4] at a step of 0.1. Higher $\delta$ means more nodes will be dropped based on their pixel discrepancy. 
In Figure~\ref{fig:ev} (b), we find that moderate node dropout (\ie, $0.1\sim 0.3$) boost the performance on DAVIS, with $\delta=0.2$ peaking the result. It demonstrates that the technique can tackle ``common-fate'' nodes, helping the model focus on informative contents.

\noindent \textbf{Path length.} In Figure~\ref{fig:ev} (c), we explore the effect of path length during training. Using clips of length 2, 4, 6, 10, we obtain paths of length 4, 8, 12, 20 for training. We see that longer sequences can improve results on DAVIS. This observation is similar to previous work~\cite{jabri2020walk}, as model can see longer views of instances for contrastive learning.
\begin{table}[!t]
\caption{\textbf{Component analysis} of our model on DAVIS benchmark.  }
	\label{tab:abl}
	\centering
	\resizebox{55mm}{!}{
\begin{tabular}{cccc}
\hline
\multicolumn{1}{c}{$G_s$} & $G_r$ & Node Dropout & $\mathcal{J}$\&$\mathcal{F}_m$    \\ \hline
$\checkmark$   &                      &              & 65.6 \\
$\checkmark$   &    $\checkmark$      &              & 67.8 (+2.2\%) \\
$\checkmark$   &     $\checkmark$     &  $\checkmark$    & 68.7 (+3.1\%)\\
\hline
\end{tabular}
}
\end{table}

\noindent \textbf{Component analysis.} We analyze the key components of our model in Table~\ref{tab:abl}. The similarity graph $G_s$ alone yields unsatisfactory results due to the puzzling negative samples. By modelling neighbor relations in $G_r$, the performance is greatly improved by 2.2\% in $\mathcal{J}$\&$\mathcal{F}_m$. Further using node dropout on the joint graph peaks the performance.

\section{Conclusion}
In this work, we present a novel self-supervised approach for learning  correspondence from unlabeled videos. Our key idea is to explore the structures and dynamics of objects from both neighboring and temporal views. To achieve this, we learn to walk on a joint space-time graph that connects nodes with neighbor relations and similarity relations. 
The superiority of our learned representation is demonstrated on three video propagation tasks.
Without fine-tuning, our method  outperforms the state-of-the-art self-supervised methods, as well as some strong fully supervised models that are designed for specific tasks. In the future, we plan to handle those  ``extremely similar instances'' --- whose correspondences are hard to find based on visual similarity --- by leveraging the motion patterns~\cite{tokmakov2017learning} or depth information~\cite{zhou2017unsupervised} from large-scale unlabeled videos. 

\section*{Acknowledgement}
This work was supported by Hong Kong Research Grants Council with Project No. CUHK 14201620.

{\small
\bibliographystyle{ieee_fullname}
\bibliography{egbib}
}

\end{document}